# Spatial-Spectral Fusion by Combining Deep Learning and Variation Model

Huanfeng Shen, S*enior Member, IEEE*, Menghui Jiang, *Student Member, IEEE*, Jie Li, *Member, IEEE*, Qiangqiang Yuan, *Member, IEEE*, Yanchong Wei, *Student Member, IEEE*, and Liangpei Zhang, *Senior Member, IEEE*

*Abstract*— in the field of spatial-spectral fusion, the model-based method and the deep learning (DL)-based method are state-of-the-art. This paper presents a fusion method that incorporates the deep neural network into the model-based method for the most common case in the spatial-spectral fusion: PAN/multispectral (MS) fusion. Specifically, we first map the gradient of the high spatial resolution panchromatic image (HR-PAN) and the low spatial resolution multispectral image (LR-MS) to the gradient of the high spatial resolution multispectral image (HR-MS) via a deep residual convolutional neural network (CNN). Then we construct a fusion framework by the LR-MS image, the gradient prior learned from the gradient network, and the ideal fused image. Finally, an iterative optimization algorithm is used to solve the fusion model. Both quantitative and visual assessments on high-quality images from various sources demonstrate that the proposed fusion method is superior to all the mainstream algorithms included in the comparison in terms of overall fusion accuracy.

## I. INTRODUCTION

Spatial–spectral fusion [1] is an important approach in remote sensing image fusion. It is aimed at obtaining a fused image with both high spatial and spectral resolutions. PAN/MS fusion (pansharpening) is the most common case in spatial-spectral fusion, aiming at integrating the geometrical details of the high spatial resolution panchromatic image (HR-PAN) and the rich spectral information of the low spatial resolution multispectral image (LR-MS) to obtain a high spatial resolution multispectral image (HR-MS) [2].

To date, a large number of PAN/MS fusion methods have been proposed, and these methods can be generally divided into four major branches [3]: 1) the component substitution (CS)-based methods; 2) the multiresolution analysis (MRA)-based methods; 3) the variation model-based methods; and 4) the deep learning (DL)-based methods. Among them, the former two branches are regarded as traditional algorithms. The CS-based methods substitute the spatial component of the LR-MS image by the HR-PAN image [4]; the MRA-based methods extract the spatial structures from the HR-PAN and inject it into the LR-MS image [5]. Due to the lack of prior knowledge in these traditional algorithms, obvious distortions can be easily caused in the spectral domain or the spatial domain, which severely degrades the quality of the fused image.

The model-based methods regard the fusion process as an ill-posed inverse optimization problem, and construct the energy functional based on the HR-PAN image, the LR-MS images and the ideal fused image. Then iterative optimization algorithm, such as the gradient descent algorithm [6], the conjugate gradient algorithm [7], the split Bregman iteration algorithm [8], and the alternating direction method of multipliers (ADMM) algorithm [9], is used to solve the model to get the fused image.

The deep learning (DL)-based methods have been proposed in recent years, which can be regarded as another new branch of PAN/MS fusion methods. Masi et al. [10] stacked the HR-PAN image with the up-sampled LR-MS image to form an input cube, and used a CNN network to learn the mapping between the input cube and the HR-MS image. Wei et al. [11] adopted a deep residual learning to learn the mapping. Yuan et al. [12] proposed a multi-scale CNN for PAN/MS fusion each layer was constituted by filters with different sizes for multi-scale features.

The two kinds of methods have their respective merits and drawbacks. On the one hand, the model-based method is flexible to handle different fusion tasks. For example, TSSC [13] can handle fusion tasks of images from the QuickBird sensor, the IKONOS sensor and the WorldView-2 sensor. However, the DL-based methods have to learn a different model for each sensor. With the sacrifice of flexibility, the DL-based tend to deliver a promising performance with the specific capability of feature extraction and learning [14]. On the other hand, due to the computational complexity, the model-based methods are usually time-consuming, while the DL-based methods is more efficient under a learned network.

As is showed above, the model-based method and the DL-based method are partly complementary. In this paper, we propose a PAN/MS fusion mothed that incorporates a deep residual gradient CNN into the model-based fusion method. Firstly, we use the deep residual gradient CNN to generate the gradient information of the HR-MS image. Then the gradient prior is plugged in a model-based optimization method, which can simultaneously utilize the flexibility of the model-based method and the feature learning capacity of deep learning to makes a high-quality fusion performance.

The rest of this paper is organized as follows. Section II gives a detailed description of the proposed method. In Section III, experiments and discussions are presented. Conclusion and future research directions are drawn in Section IV.

## II. PROPOSED METHOD

### A. The Model-Based Fusion Framework

Generally, the energy functional of model-based fusion methods in PAN/MS fusion can be represented as the following expression:

$$E(X) = f_{spectral}(X, Y) + f_{spatial}(X, Z) + f_{prior}(X) \quad (1)$$

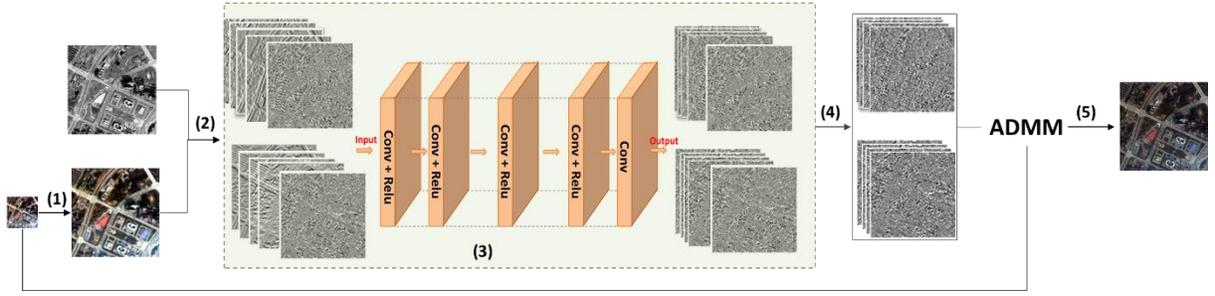

Fig. 1. Flowchart of the proposed method. (1) is the up-sample process of the LR-MS image. (2) is to obtain the horizontal and vertical gradients of the HR-PAN image and the up-sampled LR-MS image. (3) is the deep residual gradient CNN, whose output is the gradients of the HR-PAN image and the HR-MS image. (4) is to select the gradient of the HR-MS image from the output of the gradient network; (5) is constructing the fusion model and using ADMM to solve it.

where $X \in \mathbb{R}^{MN \times S}$ denotes the ideal fused image; M, N, S is the width, the height, the band number of the ideal image; $Y \in \mathbb{R}^{mn \times S}$ denotes the LR-MS image; $Z \in \mathbb{R}^{MN \times 1}$ denotes the HR-PAN image; M/m is the spatial resolution ratio of the LR-MS image to the HR-MS image The first term is the spectral fidelity model to represent the relation between the ideal fused image and the LR-MS image. The second term is the spatial enhancement model, which relates the ideal fused image to the HR-PAN image. The last term is the regularization term that imposes constraints on the ideal fused image, a Laplacian prior [15], a Huber-Markov prior [16], a total variation (TV) prior [17], a nonlocal prior [18], or a low-rank priors [19] has been proposed in many model based PAN/MS fusion methods.

### B. Deep Learning Prior for Fusion

Existing papers that combined the deep learning and the model-based optimization mainly train a set of fast and effective the regularization term, like the denoisers [14,20]. However, the spatial-spectral fusion focus on mining and integrating the high resolution information into the desired image, meaning that this enables a deep learning based spatial enhancement term into model-based methods. In general, the spatial enhancement model is constructed based on two assumptions. The first assumes a spectral degradation between the HR-MS image and the HR-PAN image, i.e., the wide band of the HR-PAN image is assumed to be a linear combination of the narrow bands of the HR-MS image [16,21]. The second assumes that the spatial structures of the ideal fused image are approximately consistent with the HR-PAN image [22], including gradient features [22,23], wavelet coefficients [24] and so on. In the first assumption, a complex nonlinear relationship existed between the HR-PAN and the spectral bands of the HR-MS image [1], which is unsuitable to express as a linear function. In the second assumption, a consistency constraint fidelity term is used to retain the high spatial information like gradient features. The relation between gradients of the HR-PAN and the HR-MS can also be nonlinear since the gradients are usually obtained by linear operators on images. Therefore, nonlinearity can also be more suitable to describe the relation of images's spatial structures.

With the capability in feature extraction and mapping learning, deep learning just have great potential to model the complex nonlinear relationship [25]. Therefore, it is more attractive to integrate the deep learning into the spatial enhancement model. In this paper, we choose the gradient features to represent the spatial structures and train mapping from the gradient of the HR-PAN images and the LR-MS images to that of the HR-MS image via a deep residual CNN.

The proposed fusion method incorporates the deep residual gradient CNN into the model-based framework. Specifically, the method mainly includes three steps. Firstly, we train a deep residual gradient CNN to obtain the gradient information of the HS-MS image, which is represented as $G$ later. Then, a model-based optimization is constructed using the LR-MS image, the gradient prior and the ideal fused image. Finally, an iterative optimization algorithm is used to solve the fusion model.

### C. Deep Residual Gradient CNN

**1) Gradient generation using deep residual CNN:** Instead of directly mapping the HR-PAN image and the LR-MS image to the HR-MS image, we stack the gradient of the HR-PAN image and that of the up-sampled LR-MS image in the spectral dimension to form an input cube. On the one hand, using the gradient directly makes the goal of network more clear, which is to obtain accurate spatial structures of the HR-MS image, i.e. the gradient. On the other hand, most pixel values in gradient image would be very close to zero, and the spatial distribution of the feature maps should be very sparse, which can transfer the gradient descent process to a much smoother hyper-surface of loss to the filtering parameters just like the residual learning[11].

Instead of use the gradient of the ground truth HS-MS image, we map the input gradient cube to the residual gradient cube. The Mean Square Error (MSE) is used as loss function.

$$loss = \frac{1}{N}\sum_{k=1}^{N}\|f(In_k)-(Tr_k-In_k)\|_F^2 \qquad (2)$$

where $In$ is the input gradient, $Tr$ is the ground truth gradient, $f(In)$ is the output of the CNN, $f(\cdot)$ means the mapping process, $N$ represents training image (patch) pairs.

Combined with the gradient, the residual learning further improves the sparsity of the network, which can not only speed up the training but also boost the leaning performance. The same size between the output and the input is convenient for mapping the residual.

**2) Architecture of the deep residual gradient CNN:** We adopt a general CNN architecture, see Fig.1. Inspired by the deep residual network for image denoising [26], the paper presents a specific improvement of the architecture to the task of gradient training. The CNN has 17 blocks, which consists of the following three types.

a) Conv+ReLU: In the first block, we use 64 filters of size 3×3×S to produce 64 feature maps, where S represents the number of spectral bands.

b) Conv+ReLU: For blocks 2~16, we use 64 filters of size 3 × 3 × 64.

c) Conv: For the last block, S filters of size 3 × 3 × 64 are exploited to produce the output. We use rectified linear units (ReLU) as activation function.

### D. The learned Gradient Guidance based Fusion Model

The energy functional used in the proposed method can be written as:

$$X = \arg\min_X \frac{1}{2}\|Y - HX\|_F^2 + \frac{\lambda_1}{2}\sum_{j=1}^{2}\|\nabla_j X - G_j\|_F^2 + \frac{\lambda_2}{2}\{\|DX\|_F^2\} \quad (3)$$

In the first term, $H \in \mathbb{R}^{mn \times MN}$ is the downsampling and blurring matrix, the spectral fidelity model is constructed based on the assumption that the observed LR-MS image can be obtained by blurring, downsampling, and the noise operators performed on the HR-MS image. In the second term, $\nabla_j \in \mathbb{R}^{MN \times MN}$ with $j = 1, 2$ means the global first-order finite difference matrix in horizontal and vertical directions respectively. $G_1 \in \mathbb{R}^{MN \times S}$ and $G_2 \in \mathbb{R}^{MN \times S}$ are the horizontal and vertical gradient image learned from the gradient network. It is assumed that the gradient of the ideal image is consistent with $G_j$. The third term is the common Laplacian prior model, where $D \in \mathbb{R}^{MN \times MN}$ indicates the Laplacian matrix. $\lambda_1$ and $\lambda_2$ are adjustable parameters used to balance the relative contribution of the three terms.

This fusion model is difficult to solve due to its large dimension. To lower the difficulty, the variable splitting technique can be used to decouple the fidelity terms and regularization term. More specifically, we adopt the alternating direction method of multipliers (ADMM) algorithm [9] to solve the model, which can be replaced by other iterative optimization algorithms.

### III. EXPERIMENTS AND DISCUSSION

To verify the effectiveness of the proposed method, both simulated and real-data experiments are performed, as described below. The proposed method is compared with five mainstream algorithms from different branches: the generalized Laplacian pyramid with modulation transfer function matched filter (MTF-GLP) [5], pansharpening with a guided filter based on three-layer decomposition (3-gui-filter) [27] and the two-step sparse coding model (TSSC) [13], the deep residual pansharpening network (DRPNN) [11], and the DNCNN-based pansharpening. In order to eliminate the effect of the different network structure, the experimenter designed DNCNN-based pansharpening, which has the similar network structure as the gradient network in the proposed fusion method.

To quantify the accuracy of fusion result, four indexes are used in the paper. They are the relative dimensionless global error in synthesis (ERGAS), the spectral angle mapper (SAM), the Q metric, and the peak-signal-noise-ratio (PSNR).

**1) Parameter setting and network training:** In (3), $\lambda_1$ and $\lambda_2$ are adjustable parameters used to balance the relative contribution of three terms We set $\lambda_1 = 0.5$, $\lambda_2 = 0.01$ after a series of experiments.

Two Gradient CNN models are trained for the QuickBird sensor and the WorldView-2 sensor respectively. When train the QuickBird Gradient CNN model, we use QuickBird images to get 8960 patches for testing and 102400 patches for training. The size of each patch is 40×40×10. When train the WorldView-2 gradient CNN model, we use WorldView-2 images to obtain 3840 patches for testing and 51200 patches for training. The size of the patches is 40×40×18 and the batchsize used in both models is 128.

**2) Test data sets:** Four data sets are employed in the simulated and real-data experiments, as follows. The gray values of each image are all normalized to [0,1].

a) The first data set is the QuickBird images, which are cropped to 250×250×4 to get the LR-MS images and 1000×1000 to get the HR-PAN images respectively. There are 12 pairs of images of different texture; the images are used in the simulated-data experiments.

b) The second data set is the WorldView-2 images, which are cropped to 250×250×8 to get the LR-MS images and 1000×1000 to get the HR-PAN images respectively. There are 6 pairs of images of different texture, the images are used in the simulated-data experiments.

c) The third data set is a pair of IKONOS images with the size of 200×200×4 and 800×800. It is used in the real-data experiment.

d) The fourth data set is a pair of WorldView-2 images with the size of 200×200×8 and 800×800. It is used in the real-data experiment.

### A. Simulated Experiments

In the simulation experiments, we first down sample the HR-PAN image and the LR-MS image to get the low resolution PAN image (LR-PAN) and the lower resolution MS image, then fuse the LR-PAN image and the lower resolution MS image to get the fused image. The original LR-MS image works as a reference to evaluate the fused image qualitatively and quantitatively.

Table I and Table II show the simulated QuickBird, WorldView-2 experimental results with the average of 12 and 6 groups, respectively. In these experiments, our SAM values and Q values are very close to the best results, but our other indicators are higher than that of other methods. Therefore, it can be demonstrated that our method can provide a better tradeoff between the spectral information fidelity and the spatial detail enhancement.

A group of simulated fusion results is selected to be displayed as true-color images in Figs. 2 and 3. By comparing the results, it can be observed that by traditional method and model-based method, sharpened spatial features are achieved, but with severe spectral distortion such as the vegetable area shown in Fig. 2(c)–(e). For the two CNN-based methods, both of them the have good performance in spectral fidelity, but are poor in spatial texture information enhancement, such as the zoom area in Fig. 2(f)–(g), Overall, the DRPNN performs better than the DNCNN-based pansharpening. The proposed method combines the advantages of model-based method and deep learning method, the fused result are the closest to the ground truth both in the fusion of the spatial details and in the preservation of spectral fidelity (see Fig. 2(h)).

Table I Quantitative results of the simulated QuickBird images (12groups)

| Method\index | ERGAS(↓) | SAM(↓) | Q(↑) | PSNR(↑) |
|---|---|---|---|---|
| MTF-GLP | 3.6042 | 3.7258 | 0.8835 | 33.5235 |
| 3-gui-filter | 3.2189 | 3.7093 | 0.8925 | 33.7267 |
| TSSC | 3.2816 | 3.4036 | 0.9085 | 34.1196 |
| DNCNN | 3.0136 | 2.9402 | 0.9144 | 34.7869 |
| DRPNN | 2.7863 | **2.6903** | 0.9265 | 35.4484 |
| Proposed | **2.4598** | 2.7124 | **0.9441** | **36.4841** |

Table II Quantitative results of the simulated WorldView-2 images (6groups)

| Method\index | ERGAS(↓) | SAM(↓) | Q(↑) | PSNR(↑) |
|---|---|---|---|---|
| MTF-GLP | 2.4655 | 3.3442 | 0.9132 | 34.1451 |
| 3-gui-filter | 2.1720 | 2.9635 | 0.9243 | 35.3450 |
| TSSC | 2.0803 | 3.0405 | 0.9276 | 35.7345 |
| DNCNN | 2.0757 | 2.9492 | 0.9303 | 35.7703 |
| DRPNN | 1.9990 | 2.8259 | **0.9384** | 35.9148 |
| Proposed | **1.8615** | **2.7028** | 0.9381 | **36.6352** |

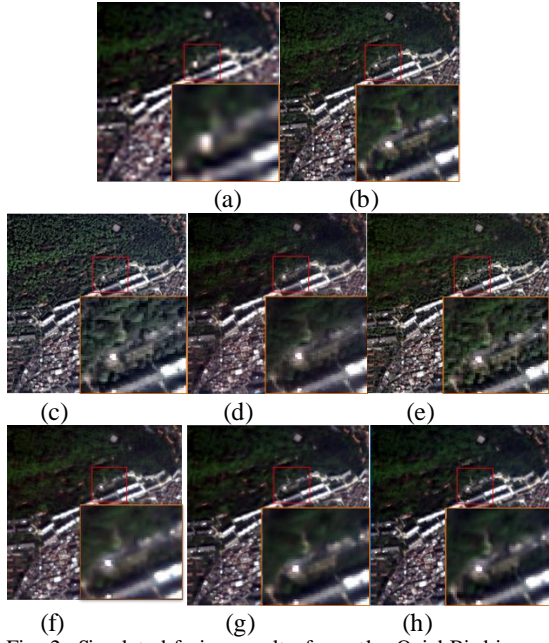

Fig. 2. Simulated fusion results from the QuickBird image. (a) Low-resolution MS image simulated by down-sampling. (b) Ground truth. (c) MTF-GLP. (d) 3-gui-filter (e) TSSC. (f) DNCNN. (g) DRPNN. (h) Proposed.

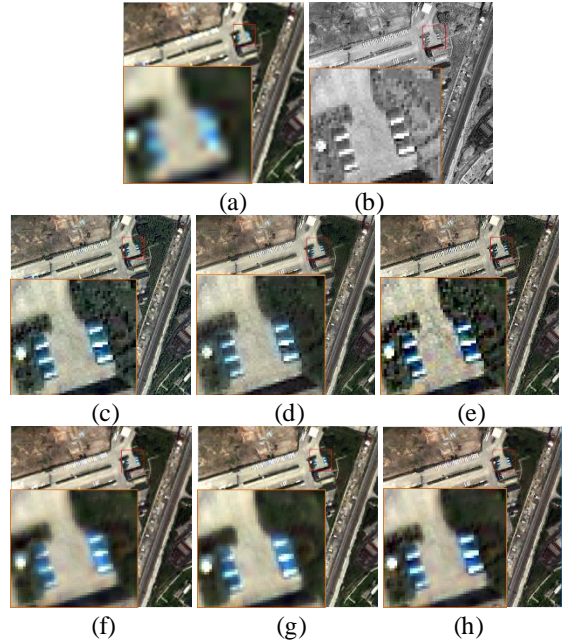

Fig. 4. Real-data fusion results from the IKONOS image. (a) the LR-MS image (b) the HR-PAN image. (c) MTF-GLP. (d) 3-gui-filter (e) TSSC. (f) DNCNN. (g) DRPNN. (h) Proposed.

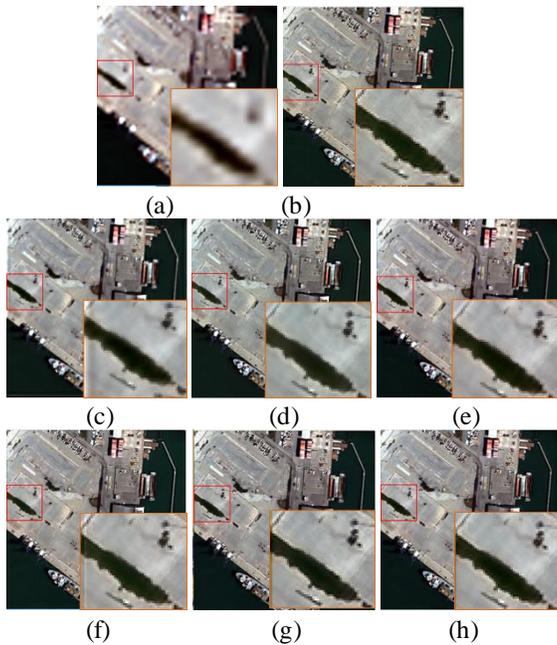

Fig. 3. Simulated fusion results from the WorldView-2 image. (a) the LR-MS image (b) Ground truth. (c) MTF-GLP. (d) 3-gui-filter (e) TSSC. (f) DNCNN. (g) DRPNN. (h) Proposed.

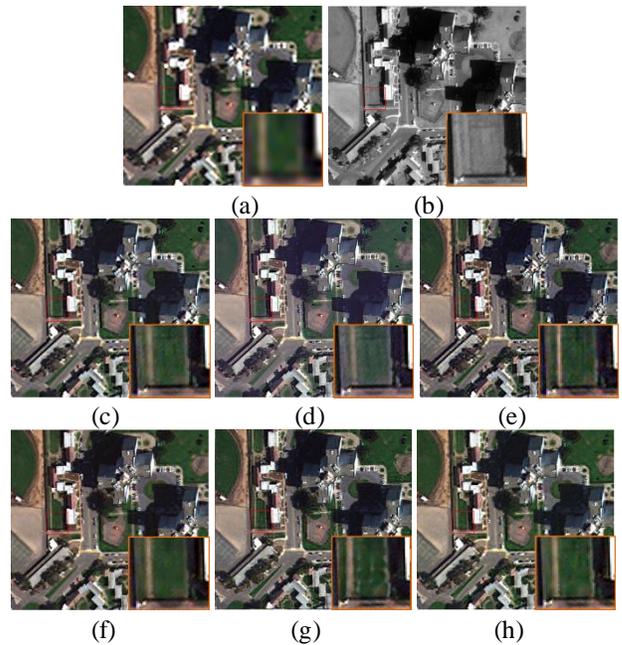

Fig. 5. Real-data fusion results from the WorldView-2 image. (a) the LR-MS image (b) the HR-PAN image. (c) MTF-GLP. (d) 3-gui-filter (e) TSSC. (f) DNCNN. (g) DRPNN. (h) Proposed.

To illustrate the superiority of the proposed method roundly, we add simulated experiments to compare the strategy of plugging the gradient network to the model-based method and that of integrating the image fusion network into the model-based method. DNCNN-model and DRPNN-model represent strategies that add the gradient of the fusion results obtained by DNCNN and DRPNN to the constructed model respectively. Compared with experimental results of Table II, the ERGAS value of DNCNN-model is 2.0518, DRPNN-model is 1.9399, and the proposed method is 1.8615. The SAM value of DNCNN-model is 2.9628, DRPNN-model is 2.7635, and the proposed method is 2.7028. The reason could be that, the purpose of the network is to obtain the accurate HR-MS image's gradient information, a lot of non-gradient information existing in image fusion network restrict the training of image's gradient, while, the gradient network better preserves the gradient of images.

### B. Real Data Experiments

To further verify the effectiveness of the proposed method, two real data sets were employed in our real-data experiments. In the real-data experiments, we fuse the HR-PAN image and the LR-MS image directly to get the fused image.

Figs. 4 and Figs. 5 show the results of IKONOS images and WorldView-2 images fused by various mainstream methods respectively. Note that due to the lack of enough IKONOS images used for training a model, the three DL-based methods used in this paper do not train a model for the IKONOS sensor, we use the model trained for the QuickBird sensor to fuse the IKONOS images instead. The real-data experiments show the same tendency as the simulated experiments.

### IV. CONCLUSION

In this paper, we have proposed a PAN/MS fusion method that incorporates the deep residual gradient CNN into the model-based framework. In the proposed method, we train a gradient network to obtain the gradient of the HR-MS image, and then utilize the learned gradient prior to construct a fusion model. Experiments on QuickBird, WorldView-2 and IKONOS datasets show the effectiveness of the proposed method.

In the future works, the proposed method can be extended in two directions. On one hand, we just utilize a simple CNN to learn the gradient information, and therefore, it is necessary to discover the CNN with different architectures, which may improve the performance. On the other hand, it is of great significance to think about other ways to integrate deep learning into the model-based methods to fully utilize the flexibility of the model-based method and the feature learning capacity of deep learning and get better results.